\documentclass[a4paper, 10pt, conference]{ieeeconf}      

\IEEEoverridecommandlockouts                              

\usepackage{cite}
\usepackage{amsmath,amssymb,amsfonts}
\usepackage{algorithmic}
\usepackage{graphicx}
\usepackage[T1]{fontenc}
\usepackage{textcomp}
\usepackage{adjustbox}
\usepackage{multirow}
\usepackage{xcolor}
\usepackage{subcaption}

\usepackage{geometry}
\usepackage{lipsum} 
\def\BibTeX{{\rm B\kern-.05em{\sc i\kern-.025em b}\kern-.08em
    T\kern-.1667em\lower.7ex\hbox{E}\kern-.125emX}}

\newgeometry{
  a4paper,
  top=25.4mm,
  bottom=36.7mm,
  left=19.1mm,
  right=13.1mm
}

\begin{document}

\title{\LARGE \bf
Unsupervised Online Detection of Pipe Blockages and Leakages in Water Distribution Networks
}

\author{
Jin Li$^{1,2}$, Kleanthis Malialis$^{1}$, Stelios G. Vrachimis$^{1}$, and Marios M. Polycarpou$^{1,2}$%
\thanks{$^{1}$KIOS Research and Innovation Center of Excellence, University of Cyprus, Nicosia, Cyprus}%
\thanks{$^{2}$Department of Electrical and Computer Engineering, University of Cyprus, Nicosia, Cyprus}%
\thanks{li.jin, malialis.kleanthis, vrachimis.stelios, mpolycar@ucy.ac.cy}%
\thanks{ORCID: \{0000-0002-3534-524X, 0000-0003-3432-7434, 0000-0001-8862-5205, 0000-0001-6495-9171\}}%
\thanks{This paper was supported by the European Research Council (ERC) under grant agreement No 951424 (Water-Futures), the European Union’s Horizon 2020
research and innovation programme under grant agreement No 739551 (KIOS CoE), and the Republic of Cyprus through the Deputy Ministry of Research, Innovation and Digital Policy.}
}

\maketitle

\begin{abstract}
Water Distribution Networks (WDNs), critical to public well-being and economic stability, face challenges such as pipe blockages and background leakages, exacerbated by operational constraints such as data non-stationarity and limited labeled data. This paper proposes an unsupervised, online learning framework that aims to detect two types of faults in WDNs: pipe blockages, modeled as collective anomalies, and background leakages, modeled as concept drift. Our approach combines a Long Short-Term Memory Variational Autoencoder (LSTM-VAE) with a dual drift detection mechanism, enabling robust detection and adaptation under non-stationary conditions. Its lightweight, memory-efficient design enables real-time, edge-level monitoring. Experiments on two realistic WDNs show that the proposed approach consistently outperforms strong baselines in detecting anomalies and adapting to recurrent drift, demonstrating its effectiveness in unsupervised event detection for dynamic WDN environments.

\end{abstract}
\begin{keywords}
water distribution networks, water leakage, pipe blockage, anomaly detection, concept drift, autoencoders
\end{keywords}

\section{Introduction}
In the face of climate change, drinking water scarcity is expected to worsen. A reliable Water Distribution Network (WDN)\cite{eliades2024smart} requires automated monitoring for early event detection to ensure supply consistency. Data-driven event detection methods that use pressure sensors, favored for their low cost, present challenges due to unlabeled events and fluctuating operating conditions, such as pipe cracks (background leakages). On the other hand, faults like pipe blockages, due to debris or valve malfunctions, significantly impact pressure. In other words, background leakages in WDNs cause minor pressure shifts, while pipe blockages trigger significant changes.

The vast majority of existing data-driven methods treat the problem of event detection as anomaly detection, and, typically, assume the existence of only one type of anomaly, for example, only pipe blockage or only leakages, but not both\cite{yuhan2021rapid, al2020review}. Furthermore, most anomaly detection methods rely on signatures or labeled data (i.e., the ground truth), which are difficult to obtain in real time for WDNs.

If a group of related data exhibits anomalies relative to the entire dataset, it is termed collective anomalies. Individually, these data instances may not be considered anomalies, but their occurrence together as a collection is deemed anomalous\cite{chandola2009anomaly}. For example, the pressure values of nodes during a pipe blockage can be considered as collective anomalies, as these values partially fall within the normal range.

This study focuses on the simultaneous detection of changes in operating conditions and anomalies in WDN. We introduce a new approach aimed at addressing the challenges associated with the detection of these events. The key contributions of this work are as follows:
\begin{enumerate}

\item We propose an unsupervised online framework for detecting pipe blockages (anomalies) and water leakages (concept drift) in water networks, combining LSTM Variational AutoEncoder (LSTM-VAE) with dual drift detection, and enabling lightweight edge deployment for real-time, distributed fault monitoring.

    
\item We evaluate two realistic WDNs and demonstrate that our method effectively detects pipe blockages amid background leakages, outperforming strong baselines and state-of-the-art approaches.


\end{enumerate}

The paper is organized as follows: Sec.~\ref{sec:related} reviews related work; Sec.~\ref{sec:method} outlines the problem and method; Sec.~\ref{sec:exp_setup} and Sec.~\ref{sec:exp_results} cover the experimental setup and results; Sec.~\ref{sec:conclusion} discusses and concludes and outlines future work.

\section{Related Work}\label{sec:related}

\subsection{Concept drift adaptation}

Data nonstationarity poses a significant challenge in certain streaming applications, often stemming from concept drift, which signifies a change in the underlying joint probability distribution. Approaches to adapt to concept drift are often classified as passive or active\cite{ditzler2015learning}.

\textbf{Passive methods} implicitly address drift using incremental learning, which is the continuous adaptation of the model without complete re-training\cite{losing2018incremental}. In this category, methods can be categorized into memory-based and ensemble methods. Memory-based algorithms utilize a memory component to retain a set of recent examples on which the classifier is trained\cite{widmer1996learning,malialis2020online}; Ensemble methods, on the other hand, consist of a set of classifiers that can be dynamically added or removed based on their performance\cite{minku2011ddd}. 

\textbf{Active methods} rely on explicitly identifying changes in the data distribution to initiate an adaptation mechanism \cite{ditzler2015learning}. Two primary categories of detection mechanisms are investigated: statistical tests and threshold-based mechanisms. Statistical tests monitor the statistical characteristics of the generated data, while threshold-based mechanisms observe prediction errors and compare them to a predefined threshold.

\textbf{Hybrid methods}, such as HAREBA \cite{malialis2022hybrid}, are proposed to combine the strengths of both active and passive methods. Another work \cite{li2023autoencoder} also employs an autoencoder and leverages on the advantages of both incremental learning and drift detection based on the Mann-Whitney U Test. The method proposed in this paper uses unsupervised drift detection.  

\subsection{Anomaly detection}\label{sec:ano}

Several traditional machine learning techniques have been suggested for anomaly detection, notably the Local Outlier Factor (LOF) \cite{breunig2000lof} and the Isolation Forest (iForest) algorithm \cite{liu2008isolation}. LOF gauges the local density of data points to pinpoint outliers, while iForest prioritizes isolating anomalies over characterizing normal behavior. By forming an ensemble of trees, iForest discerns anomalies by their relatively shorter average path lengths within the trees.


\textbf{Collective anomalies} are anomalous sequences where individual points may appear normal~\cite{ahmed2020deep}, requiring temporal pattern recognition. Methods like DiFF-RF \cite{marteau2021random} and~\cite{rosenberger2022extended} address them but may confuse anomalies with concept drift. The unsupervised VAE4AS~\cite{li2024unsupervised} detects anomalous sequences in non-stationary i.i.d. data. Our work builds upon VAE4AS by adapting it for WDNs, handling non-stationary time series with an LSTM-VAE, and tackling topology-dependent fault propagation.

\subsection{Anomaly detection in WDNs}

Few studies have addressed pipe blockage detection using data-driven methods. Yuhan et al.~\cite{yuhan2021rapid} propose a lightweight threshold-based approach triggered when real-time flow drops below levels estimated from historical flow and rainfall data. While simple, it requires both flow and rainfall measurements, limiting applicability in sparsely instrumented networks. In contrast, pipe burst detection has received more attention, as both bursts and blockages cause pressure changes. Relevant methods include PCA with Hotelling’s T2~\cite{palau2012burst}, an ensemble CNN using statistical features~\cite{kim2022ensemble}, and a GCN-based framework leveraging WDN structure~\cite{zanfei2022novel}.

More broadly, leak detection methods in pipeline monitoring are typically classified into two categories: those relying on directly measurable quantities (e.g., inflows, outflows, pressures, temperatures) and those based on non-measurable internal states or model parameters, which require modeling and estimation techniques~\cite{al2020review}. Comprehensive reviews of these approaches can be found in~\cite{islam2022review, al2020review}.

\textbf{Research gap}. The previously mentioned methods rely on the assumption that only a single type of anomalous event occurs, while the rest of the network operates under stationary background conditions. However, they do not account for the coexistence of faults with potential background leakages, which is the focus of this work. Considering the pressure changes induced by these events, we propose an online, unsupervised method that detects pipe blockages as collective anomalies and simultaneously detects and adapts to background leakage-induced variations as concept drift, setting it apart from existing approaches.


\section{Problem Formulation and Proposed Method}\label{sec:method}

WDNs are modeled as graphs, where nodes represent junctions and undirected edges represent pipes. A graph $G = (V, E)$ consists of node set $V$ and edge set $E$, with each edge $e_{i,j} = (v_i, v_j) \in E$, $i \ne j$. Pressure sensors are commonly used due to their low cost and ease of deployment. As full sensor coverage is impractical, only $N < |V|$ sensors are installed. At time $t$, the input signals from all sensors are represented as $x^t \in R^N = \{x^t_1, \dots, x^t_N\}$, where $x^t_i$ corresponds to the measurements from sensor $i$. 
Fig.~\ref{fig:formulation} shows a simplified example with four sensors placed on six nodes.

This study detects pipe blockages and background leakage using pressure sensor data, formulating them as anomaly detection (AD) and drift detection (DD) tasks, respectively. Each sensor $i$ is equipped with its own detectors $(AD_i, DD_i)$, as illustrated by the orange boxes in Fig.~\ref{fig:formulation}, with no communication between sensors, enabling fully decentralized edge deployment for scalable and low-latency fault detection. At each time step $t$, each detector only accesses its local reading $x^t_i$. Node-level detection mechanisms for AD and DD are detailed in the following sections.

\begin{figure}[!t]
	\centering	
    \includegraphics[scale=0.35]{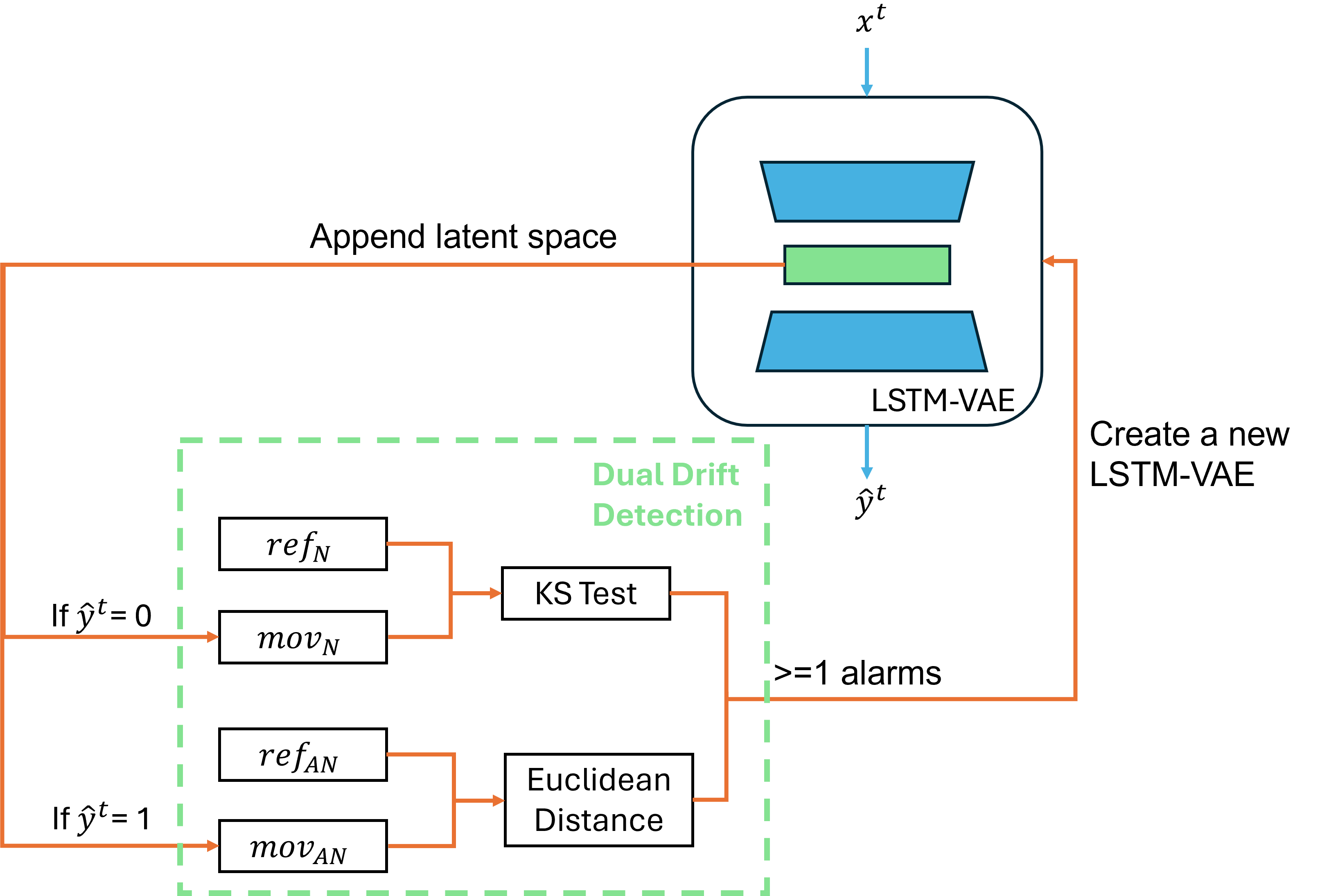}
	\caption{Overview of the proposed method.}
	\label{fig:method}
\end{figure}

The overview of the proposed method design is shown in Fig.~\ref{fig:method}, which corresponds to the orange box in Fig.~\ref{fig:formulation}. To simplify equation notation, we use $x^t, y^t$ instead of $x^t_i, y^t_i$ to represent the input and its corresponding label for each arriving instance at time $t$. The prediction part is displayed in blue. The system first observes the instance $x^t$ at time $t$, and the LSTM-VAE-based model outputs a prediction $\hat{y}^t \in \{0,1\}$. If the instance is classified as normal ($\hat{y}^t=0$), its encoding is added to $mov_{N}$ for statistical testing. If the instance is classified as anomalous ($\hat{y}^t=1$), anomalous instances' encodings are appended to sliding windows $mov_{AN}$ for distance-based testing. In cases of drift presence, normal instances affected by drift may be classified as normal or anomalous by the current classifier, depending on the similarity between drifted data and normal data. Considering this, two drift detection (DD) methods, DD1 and DD2, are introduced, with details provided in a subsequent section. Upon the activation of an alarm flag by any DD, a new LSTM-VAE model is instantiated and trained accordingly.


\subsection{Detecting pipe blockages}

In this study, we view the problem of detecting pipe blockages as an anomaly detection problem.

\textbf{Model.} 
Autoencoders are effective for anomaly detection by learning to reconstruct normal data with minimal error. However, standard VAEs assume independent and i.i.d. data, limiting their ability to model temporal dependencies in time-series data. To address this, we integrate a VAE with LSTM networks, replacing the feed-forward layers with LSTM to capture sequential patterns.

LSTM, introduced in \cite{hochreiter1997long}, mitigates vanishing gradients in recurrent networks through gating mechanisms, enabling effective modeling of long-term dependencies. By incorporating LSTM into VAE, our model learns a structured latent space that preserves temporal correlations.

A VAE models the latent distribution $q(z|x)$ as a multivariate Gaussian, regularized via Kullback-Leibler (KL) divergence. The total loss function consists of reconstruction loss and KL divergence, formulated as:
\vspace{-2pt}
\begin{equation}\label{eq:vae}
l_{VAE}(x, \hat{x}) = l_{AE}(x, \hat{x}) + \beta \cdot l_{KL}(x),
\end{equation}
\noindent where $l_{AE}(x, \hat{x})$ is the reconstruction loss between input $x$ and output $\hat{x}$, and $\beta$ is a weighting coefficient that balances the reconstruction and regularization terms. Unlike standard VAEs that reconstruct steps independently, LSTM-VAEs use past data to better detect temporal anomalies. We enhance adaptability by integrating drift detection to guide $l_{VAE}$, boosting robustness in dynamic settings.


The LSTM-VAE assumes anomalies yield higher loss than normal data. An adaptive threshold $\theta^t$, set as the maximum training loss, is used for detection. When an alarm is triggered, $\theta^t$ is updated with new data. An instance $x^{t+\Delta}$ is flagged as anomalous if its cumulative loss and that of preceding values exceed $\theta^t$. This is applied iteratively over sliding windows to generate predictions.

\subsection{Detecting water leakages} The problem of detecting water leakages is viewed as a concept drift detection problem. Specifically, we employ a dual drift detection (DD) mechanism consisting of both a statistical test (DD1) and a distance-based approach (DD2).

\textbf{DD1: Statistical test}. The KS test, a non-parametric method, avoids distributional assumptions. Given $ref_N$ and $mov_N$ containing $d$ latent layer windows, where $ref_N = \{ref_{latent_1},..., ref_{latent_d}\}$ and $mov_N = \{mov_{latent_1},..., mov_{latent_d}\}$, we apply the KS test to each dimension $i$. This test measures the maximum disparity between their cumulative distributions, providing a statistical metric \cite{nitta2023detecting}. The $p$-value is determined as:

\begin{equation}\label{eq:ks}
\begin{aligned}
p_{value} &= 2 \sum_{i=1}^{\infty}(-1)^{i-1} e^{-2 i^2 \gamma^2} \\
\text{where,} \quad 
\gamma &= \left(\sqrt{N_{\text{eff}}} + 0.12 + \frac{0.11}{\sqrt{N_{\text{eff}}}} \right) KS_{\text{dis}} \\
KS_{\text{dis}} &= \max \left| F(ref_{\text{latent}_i}) - F(mov_{\text{latent}_i}) \right|, \\
N_{\text{eff}} &= \frac{W_{\text{drift}}^2}{2 W_{\text{drift}}}
\end{aligned}
\end{equation}

Two flags, $flag_{warn}$ and $flag_{alarm}$, are established to indicate concept drift. The warning flag signals potential drift, while the alarm flag confirms actual drift, rejecting the null hypothesis ($H_0$) in favor of the alternative hypothesis ($H_1$). The thresholds satisfy $P_{warn} > P_{alarm}$.

\begin{figure}[!t]
	\centering
	\includegraphics[scale=0.4]{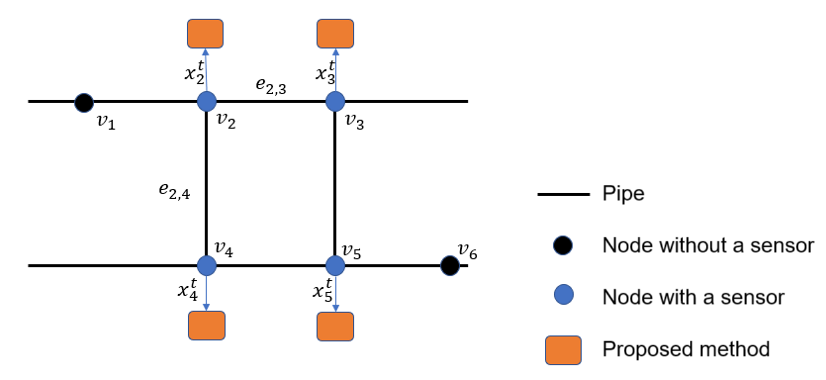}
	\caption{Placement of the proposed method in a simplified WDN example.}
	\label{fig:formulation}
\end{figure}

\begin{figure}[!t]
\begin{subfigure}{\columnwidth}
  \centering
  \includegraphics[width=.6\columnwidth]{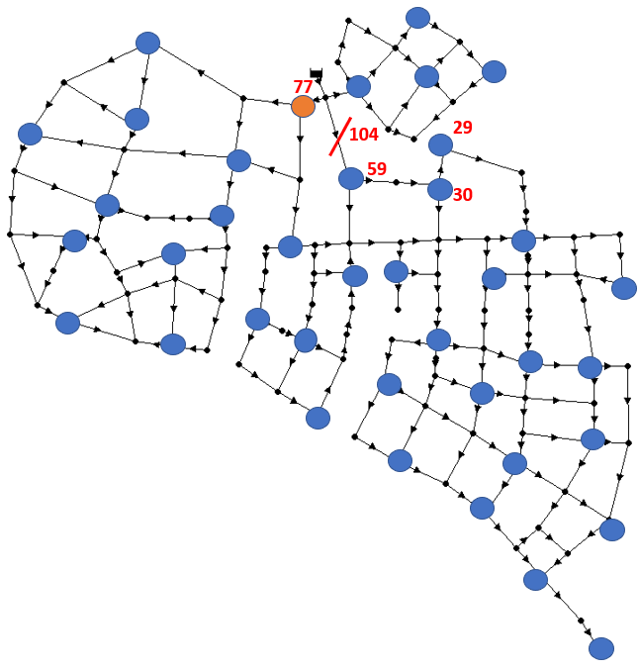}
  \caption{ZJ network}
  \label{fig:ZJ}
\end{subfigure}%

\begin{subfigure}{\columnwidth}
  \centering
  \includegraphics[width=0.9\columnwidth]{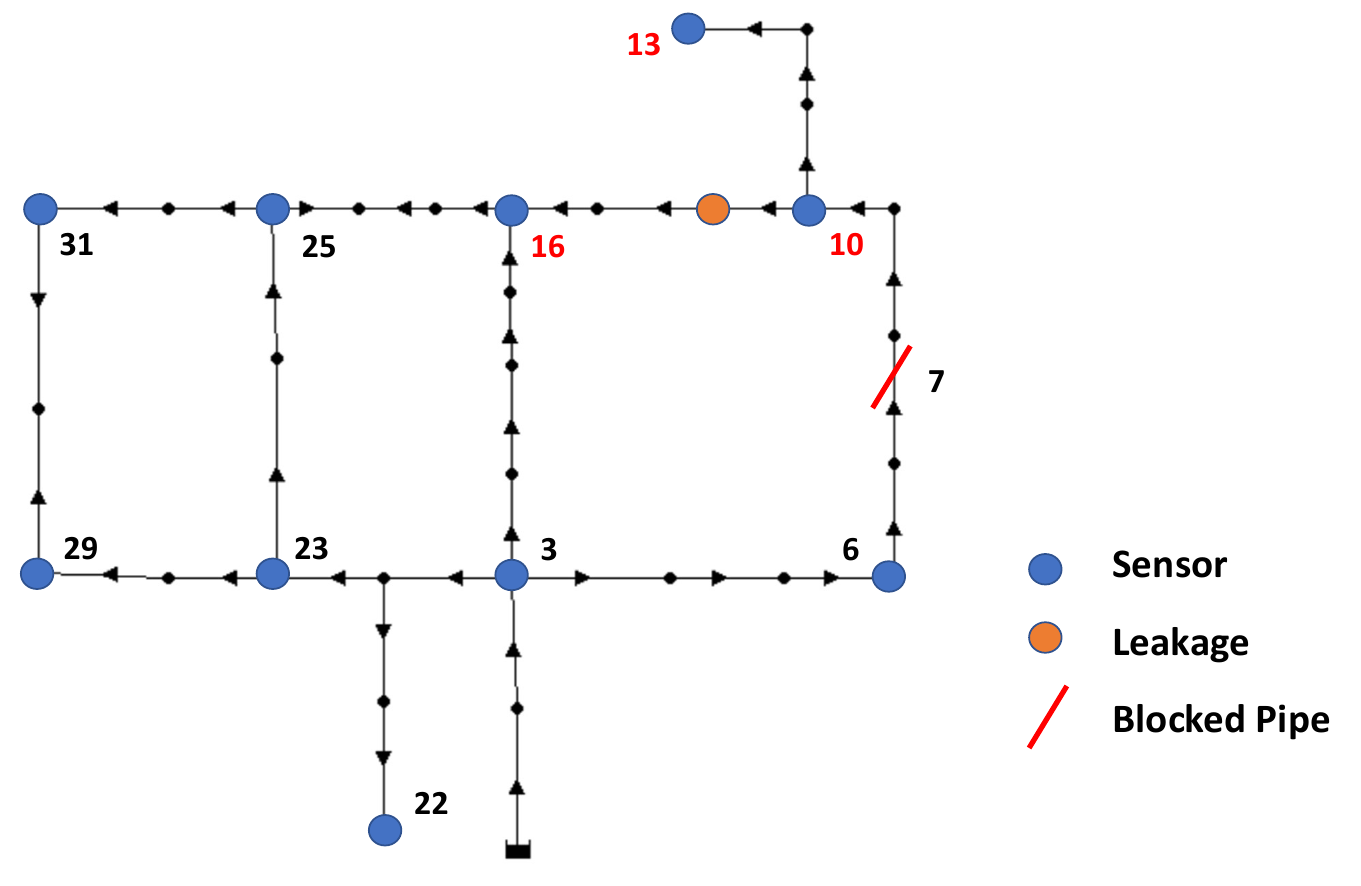}
  \caption{Hanoi network}
  \label{fig:hanoi}
\end{subfigure}
\caption{Illustration of Hanoi and ZJ network. }
\label{fig:WDN_fig}
\end{figure}

\textbf{DD2: Distance-based}. The Euclidean distance between $ref_{AN}$ and $mov_{AN}$ determines drift, as defined in Eq.~(\ref{eq:ed}) and Eq.~(\ref{eq:ed_flag}). Here, $ref_{ANij}$ and $mov_{ANij}$ denote elements at the \(i\)-th row and \(j\)-th column of $ref_{AN}$ and $mov_{AN}$, respectively. Each row represents a point in multi-dimensional space, while each column corresponds to a specific feature. The threshold $DIS_{thre}$ is set offline.

\begin{equation}\label{eq:ed}
DIS(ref_{AN}, mov_{AN})=\sqrt{\sum_{i=1}^n \sum_{j=1}^m\left(ref_{AN i j}-mov_{AN i j}\right)^2}
\end{equation}
\begin{equation}\label{eq:ed_flag}
{flag} = alarm \quad \text{if} \quad \text{DIS}(ref_{AN}, mov_{AN}) > DIS_{thre}
\end{equation}

The warning mechanism applies only to DD1. If $flag_{warn}$ is raised but not $flag_{alarm}$, instances are stored in $mov_{warn}$. To reduce false alarms, if $flag_{warn}$ persists beyond $expiry\_time$ without triggering $flag_{alarm}$, it is reset, and $mov_{warn}$ is cleared. When an alarm is activated, an autoencoder is retrained using $mov_{warn}$ or 500 post-alarm instances, depending on the drift detector. The threshold is updated accordingly, and the windows $mov_{N}$, $mov_{warn}$, and $mov_{AN}$ are cleared, resetting all flags. The reference window $ref_{N}$ is then repopulated with new normal instances.

\subsection{Computational analysis}


The method uses five memory buffers: $ref_N$ and $mov_N$ (each of size $W_{\text{drift}}$) store normal instances for drift detection during prediction via the KS test; $ref_{AN}$ and $mov_{AN}$ (each of size $W_{\text{distance}}$) store anomalies for the anomaly distance test; $mov_{warn}$ (size $W_{\text{warn}}$) buffers post-drift data for model training, which is only triggered when flag alarm is raised. At each time step, the LSTM-VAE performs lightweight forward inference to compute Eq.~\ref{eq:ed_flag}. This buffer-based design ensures low memory and compute overhead, making the approach suitable for efficient edge deployment.

\section{Experimental Setup}\label{sec:exp_setup}

\subsection{Water Distributions Networks}\label{sec:data_gen}

\noindent\textbf{Hanoi network and ZJ network.} The Hanoi network~\cite{fujiwara1990two} is a benchmark WDN with 32 nodes, 34 pipes, and one reservoir. The Zhi Jiang (ZJ) network \cite{zheng2011combined} in eastern China, represents a real WDN with 164 pipes, 113 demand nodes, 50 primary loops, and a reservoir with a fixed 45 m head. Fig.~\ref{fig:WDN_fig} provides an overview, with black arrows indicating initial water flow directions.

\subsection{Datasets}
We generate two scenarios with the Hanoi and ZJ networks, each spanning one year with 30-minute sampling, yielding 17,520 data points. Gaussian noise \(N(0,0.01)\) is added to all sensors. The dataset for pipe blockages is generated using WNTR \cite{klise2017software}, an advanced open-source tool for WDN resilience analysis. Leakage simulation follows the LeakDB method \cite{vrachimis2018leakdb}, using historical demand data. Pipe blockages are simulated by intermittently closing and opening pipes. Pressure measurements at nodes serve as key anomaly indicators. Pipe blockages occur at timesteps 2000-3000 and 8000-9000, while background leakages occur from 5000-15000. In Hanoi, pipe 7 is blocked and node 14 leaks (hole diameter = $8.9$ cm); in ZJ, pipe 104 is blocked and node 77 leaks (hole diameter = $3.0$ cm). Pressure sensors are blue; black-labeled nodes are for analysis, red-labeled ones downstream of blockages for comparison. Blockages and leakages are shown in Fig.~\ref{fig:WDN_fig}.

\noindent\textbf{Data pre-processing.} We apply Seasonal and Trend decomposition using Loess (STL) \cite{cleveland1990stl} to split one-year historical data into trend, seasonality, and residual components. For each arriving point, the corresponding trend and residual from historical data are subtracted to preserve seasonality. The STL period is set to 336 (one week). We assume that the seasonal component remains stationary over the one-year period, without accounting for potential external influences.

\subsection{Compared methods}

We compare the proposed method with three other methods, iForest++, LOF++, and VAE4AS. The first two are classical anomaly detection methods and VAE4AS is a method dealing with anomalous sequences. 

\noindent\textbf{iForest++}\cite{liu2008isolation}: A state-of-the-art anomaly detection method as described in Section~\ref{sec:ano}. To adapt to drift, incremental learning is adopted after every 1000 instances. 

\noindent\textbf{LOF++}\cite{breunig2000lof}: Another popular method to compare. Details are provided in Section~\ref{sec:ano}. To adapt to drift, incremental learning is adopted after every 1000 instances. 
   
\noindent\textbf{VAE4AS}\cite{li2024unsupervised}: In this comparison, the parameter settings remain identical to those recommended in the original paper.

\noindent\textbf{Proposed method}: As described in Section~\ref{sec:method}. For reproducibility, the hyper-parameters of the proposed method are provided in Table I. 

For VAE4AS and the proposed method, parameters are set as: $W_{warn}=1000$, $W_{drift} = 200$, $W_{distance} = 50$, $P_{alarm} = 0.0001$ and $0.001$, and $expiry\_time = 100$. Retraining epochs are 500.

\subsection{Performance metrics and evaluation method}

The geometric mean is a robust and widely accepted metric for imbalanced classification \cite{sun2006boosting}, defined as $G\text{-}mean = \displaystyle \sqrt{R^+ \times R^-}$, where $R^+ = TP / P$ is the recall of the positive class, $R^-=TN / N$ is the recall (or specificity) of the negative class. G-mean is insensitive to class imbalance and favors high, balanced recall values. We adopt prequential evaluation with a fading factor of $0.99$, which converges to Bayes error under stationarity and removes the need for a holdout set \cite{gama2013evaluating}. G-mean is computed per time step and averaged over 10 runs; error bars show standard error.

\begin{table*}[!h]
\caption{Hyper-parameter values for the proposed method}\label{tab:params_lvae}
\begin{adjustbox}{width=1.0\textwidth}

\begin{tabular}{|c|c|c|c|c|l|c|c|c|c|c|l|}
\hline
             & \begin{tabular}[c]{@{}c@{}}Learning \\ rate\end{tabular} & \begin{tabular}[c]{@{}c@{}}Hidden \\ layers\end{tabular} & \begin{tabular}[c]{@{}c@{}}Mini-batch \\ size\end{tabular} & \begin{tabular}[c]{@{}c@{}}Dropout \\ rate\end{tabular} & Timestep                & \begin{tabular}[c]{@{}c@{}}Weight \\ initializer\end{tabular} & Optimizer & \begin{tabular}[c]{@{}c@{}}Hidden \\ activation\end{tabular} & \begin{tabular}[c]{@{}c@{}}Num. \\ epochs\end{tabular} & \begin{tabular}[c]{@{}c@{}}Output \\ activation\end{tabular} & \multicolumn{1}{c|}{\begin{tabular}[c]{@{}c@{}}Loss \\ function\end{tabular}} \\ \hline
Hanoi and ZJ & 0.001                                                    & [8, 2]                                                   & 64                                                         & 0.1                                                     & \multicolumn{1}{c|}{10} & He Normal                                                     & Adam      & Leaky ReLU                                                   & 100                                                     & Softmax                                                      & Square Error                                                                  \\ \hline
\end{tabular}
\end{adjustbox}
\end{table*}

\begin{table}[!h]
\centering
\caption{Detection performance per node (True positives at downstream nodes of pipe blockages highlighted in \textcolor{red}{red})}\label{tab:detection}
\begin{adjustbox}{width=0.5\textwidth}
\begin{tabular}{|c|c|c|c|c|c|c|c|c|c|c|c|}
\hline
                                                                           &                                                          & N23  & N29  & N31  & N16 & N25  & N13  & N10  & N6 & N3 & N22 \\ \hline
\multirow{2}{*}{\begin{tabular}[c]{@{}c@{}}Pipe 7\\ blocked\end{tabular}}  & \begin{tabular}[c]{@{}c@{}}True\\ Positive\end{tabular}  & 314  & 147  & 288  & \textcolor{red}{991} & 685  & \textcolor{red}{1000} & \textcolor{red}{1000} & 0  & 0  & 182 \\ \cline{2-12} 
                                                                           & \begin{tabular}[c]{@{}c@{}}False\\ Positive\end{tabular} & 349  & 82   & 82   & 174 & 337  & 82   & 0    & 0  & 0  & 198 \\ \hline
\multirow{2}{*}{\begin{tabular}[c]{@{}c@{}}Pipe 23\\ blocked\end{tabular}} & \begin{tabular}[c]{@{}c@{}}True\\ Positive\end{tabular}  & \textcolor{red}{1000} & \textcolor{red}{1000} & \textcolor{red}{1000} & \textcolor{red}{990} & \textcolor{red}{1000} & 121  & 173  & 69 & 0  & 55  \\ \cline{2-12} 
                                                                           & \begin{tabular}[c]{@{}c@{}}False\\ Positive\end{tabular} & 349  & 82   & 82   & 174 & 337  & 82   & 0    & 0  & 0  & 198 \\ \hline
\end{tabular}
\end{adjustbox}
\end{table}

\begin{figure}[!t]
\begin{subfigure}{0.5\columnwidth}
  \centering
  \includegraphics[width=1\columnwidth]{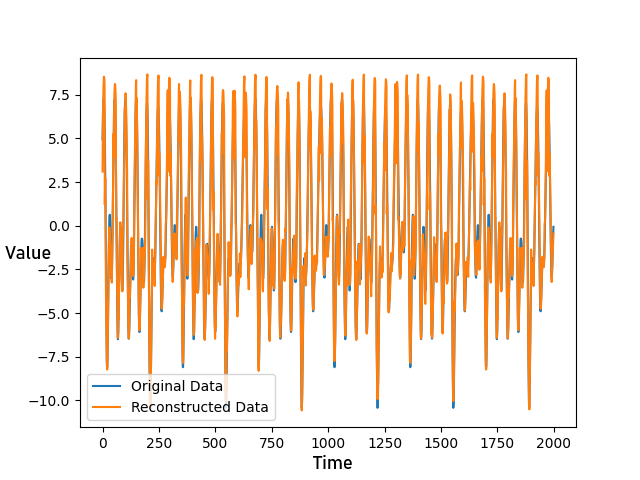}
  \caption{Reconstruction of LSTM-VAE}
  \label{fig:lvae_recon}
\end{subfigure}%
\begin{subfigure}{0.5\columnwidth}
  \centering
  \includegraphics[width=1\columnwidth]{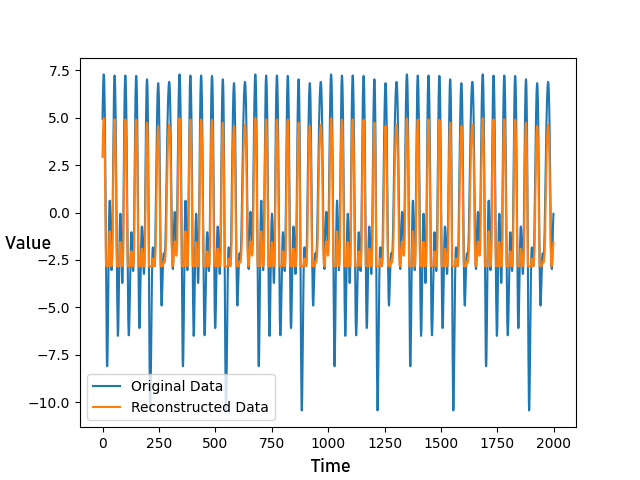}
  \caption{Reconstruction of VAE}
  \label{fig:vae_recon}
\end{subfigure}
\caption{Reconstruction of LSTM-VAE and VAE.}
\label{fig:lvae_vae}
\end{figure}

\begin{figure}[!t]
\begin{subfigure}{.5\columnwidth}
  \centering
  \includegraphics[width=0.95\columnwidth]{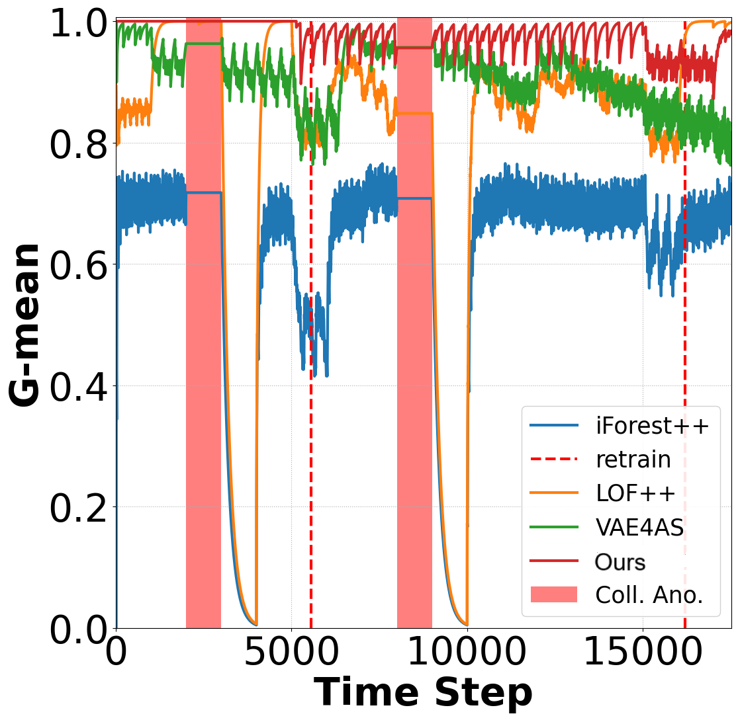}
  \caption{Node 10}
  \label{fig:node10}
\end{subfigure}%
\begin{subfigure}{.5\columnwidth}
  \centering
  \includegraphics[width=0.95\columnwidth]{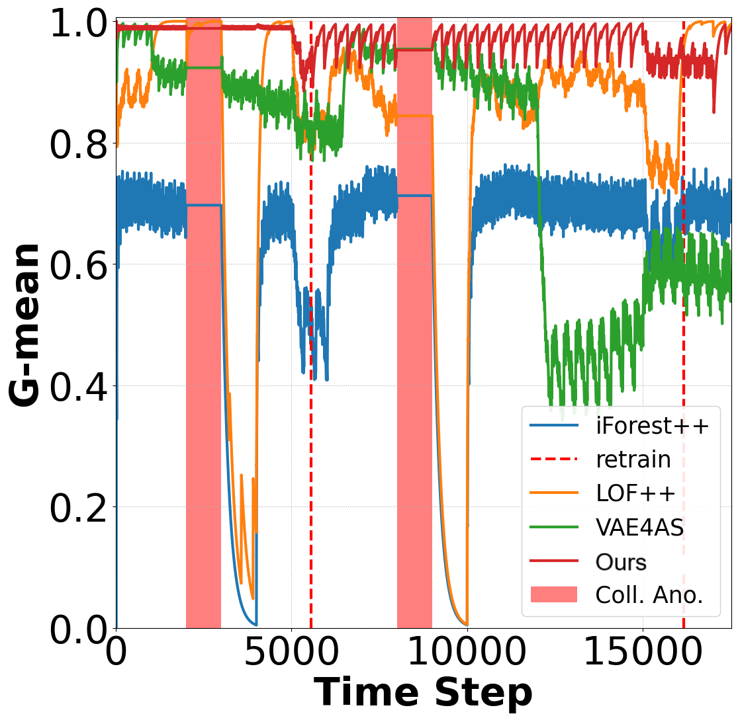}
  \caption{Node 13}
  \label{fig:node13}
\end{subfigure}
\caption{Performance of iForest++, LOF++, VAE4AS and the proposed method with the Hanoi network.}
\label{fig:hanoi_compare}
\end{figure}

\begin{figure}[!t]
\begin{subfigure}{.5\columnwidth}
  \centering
  \includegraphics[width=0.95\columnwidth]{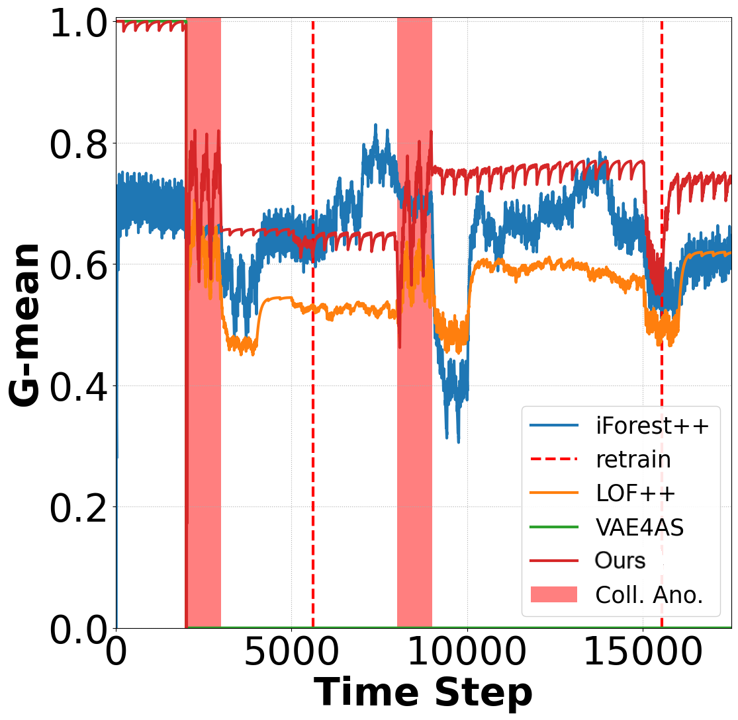}
  \caption{Node 29}
  \label{fig:node29}
\end{subfigure}%
\begin{subfigure}{.5\columnwidth}
  \centering
  \includegraphics[width=0.95\columnwidth]{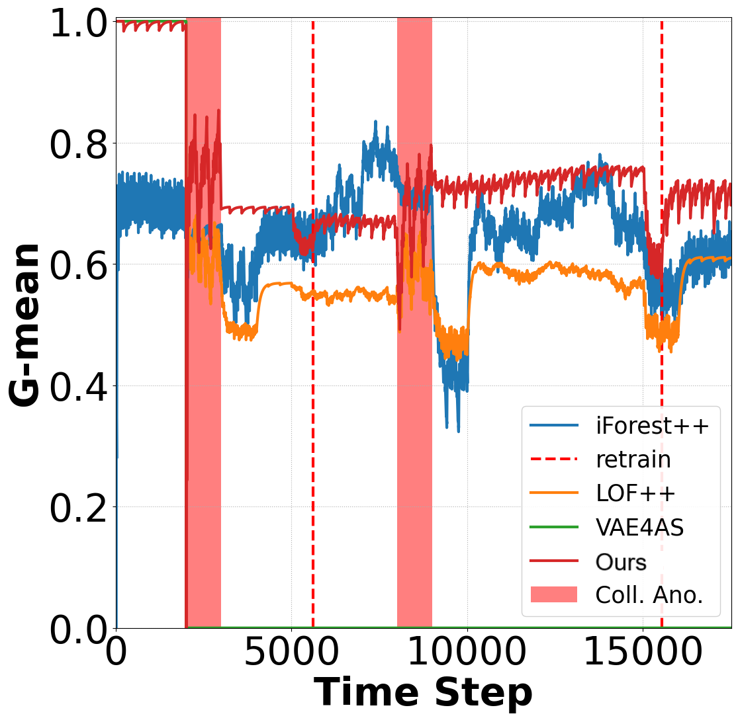}
  \caption{Node 30}
  \label{fig:node30}
\end{subfigure}
\caption{Performance of iForest++, LOF++, VAE4AS and the proposed method with the ZJ network.}
\label{fig:ZJ_compare}
\end{figure}

\section{Experimental Results}\label{sec:exp_results}

\subsection{Role of the model component}
We compare VAE and LSTM-VAE to highlight the latter’s strength in time-series modeling. Both models are trained on 5000 normal data points and validated on 2000 points from node 16 in Hanoi (outlined in Section~\ref{sec:data_gen}). In Fig.~\ref{fig:lvae_recon}, the small reconstruction loss results in overlapping curves, making differences less visible. In contrast, Fig.~\ref{fig:vae_recon} shows a larger loss. These results indicate that LSTM-VAE captures temporal dependencies more effectively than VAE, which lacks temporal modeling and yields poorer reconstructions.

\subsection{Role of the drift detector}
We evaluate the proposed method on the first 5000 instances in two Hanoi scenarios, where pipes 7 and 23 are blocked (timesteps 2000–3000) without leakage. Table~\ref{tab:detection} reports per-node results, with red-highlighted true positives corresponding to downstream nodes of the blockage, which show higher classification accuracy. Combining flow direction (Fig.~\ref{fig:formulation}) with detection results reveals more anomalies at downstream nodes. This pattern may guide blocked pipe localization. While such node-level differences suggest potential for fault isolation, we focus on overall detection performance, leaving localization to future work.

\subsection{Comparative study}

We compare iForest++, LOF++, VAE4AS, and the proposed method. Empirical analysis shows that downstream nodes of blocked pipes yield the best classification due to significant pressure differences. Thus, we select these nodes for the comparative study. Leakage (timesteps 5000-15000) represents recurrent drift, while blockages occur at 2000-3000 and 8000-9000. Collective anomalies are marked in red, with data generation details in Section~\ref{sec:data_gen}.

As shown in Fig.\ref{fig:hanoi_compare} and Fig.\ref{fig:ZJ_compare}, the proposed method outperforms the baselines. The performance at different nodes is influenced by the extent to which their pressure is affected by the blockage. In the ZJ network, due to its more complex topology—such as a higher degree of pipe interconnectivity—and the greater distance between some nodes and the blockage location, the performance (Fig.\ref{fig:ZJ_compare}) is lower compared to that in the Hanoi network (Fig.\ref{fig:hanoi_compare}). Additionally, the presence of alternative flow paths caused by pipe splitting further weakens the blockage impact on downstream nodes in the ZJ network.

Retraining points are indicated in the figures. G-mean drops at concept drift events (background leakage at 5000 and 15000) and recovers after retraining, highlighting the drift’s impact. Our method accurately detects drift without false alarms, maintaining G-mean above 0.7 on ZJ and over 0.9 on Hanoi. In contrast, VAE4AS is unstable, iForest++ fluctuates around 0.7, and LOF++ performs well on Hanoi but degrades on ZJ.

Overall, our method outperforms others, demonstrating robust anomaly detection and adaptation to concept drift in non-stationary environments.



\section{Discussion and Conclusions}\label{sec:conclusion}

We propose a label-free, data-driven method for pipe blockage detection under changing WDN conditions, where background leakages manifest as gradual pressure shifts (concept drift) and blockages cause abrupt changes (collective anomalies). Our method adopts a decentralized architecture suitable for edge deployment, with one LSTM-VAE per sensor processing local 1D time series. This design supports real-time, scalable anomaly detection close to the data source, reducing latency and communication overhead. Although resource use may rise in large systems, sparse sensor placement in real-world WDNs limits the number of models and keeps computational cost low. Detection performance depends on sensor coverage—blockages near unsensored areas may be missed. However, leveraging network topology and flow direction can help reduce this risk. Experiments on realistic Hanoi and ZJ networks show our approach outperforms strong baselines, with LSTM-VAE capturing temporal patterns and enabling cross-node anomaly detection. While the method demonstrates potential for fault isolation, further work is needed to validate its effectiveness. Additionally, beyond background leakage, future work will address long-term factors like climate change, which may alter WDN behavior.


\bibliographystyle{IEEEtran}
\bibliography{paper}

\end{document}